\documentclass[preprint,12pt]{elsarticle}

\usepackage{amssymb}
\usepackage{times}
\usepackage{soul}
\usepackage{url}
\usepackage[hidelinks]{hyperref}
\usepackage[utf8]{inputenc}
\usepackage[small]{caption}
\usepackage{graphicx}
\usepackage{amsmath}
\usepackage{amsthm}
\usepackage{booktabs}
\usepackage{algorithm}
\usepackage{algorithmic}
\usepackage{multirow}

\journal{Neurocomputing}

\begin{document}

\begin{frontmatter}

\title{EvoGAN: An Evolutionary Computation Assisted GAN}

\author[1,2,4,5]{Feng Liu\fnref{fn1}}
 
 \author[1,4]{HanYang Wang\fnref{fn1}}
 
  \author[1,4]{Jiahao Zhang}
 
  \author[3]{Ziwang Fu}
 
  \author[1,4,5]{Aimin Zhou\corref{cor1}}
 
  \author[2]{Jiayin Qi\corref{cor1}}
 
  \author[1]{Zhibin Li\corref{cor1}}

 \affiliation[1]{organization={School of Computer Science and Technology, East China Normal University}}
 \affiliation[2]{organization={Institute of Artificial Intelligence and Change Management, University of International Business and Economics, Shanghai}}
 \affiliation[3]{organization={School of Computer Science, Beijing University of Posts and Telecommunications}}
 \affiliation[4]{organization={Institute of AI Education, East China Normal University}}
 \affiliation[5]{organization={Shanghai Key Laboratory of Mental Health and Psychological Crisis Intervention, School of Psychology and Cognitive Science, East China Normal University}}

\cortext[cor1]{Corresponding author}
\fntext[fn1]{The two authors contribute equally in this work.\\
ORCID(s): 0000-0002-5289-5761 (Feng Liu);0000-0002-5227-5765(Hanyang Wang)}

\begin{abstract}
The image synthesis technique is relatively well established which can generate facial images that are indistinguishable even by human beings. However, all of these approaches uses gradients to condition the output, resulting in the outputting the same image with the same input. Also, they can only generate images with basic expression or mimic an expression instead of generating compound expression. In real life, however, human expressions are of great diversity and complexity. In this paper, we propose an evolutionary algorithm (EA) assisted GAN, named EvoGAN, to generate various compound expressions with any accurate target compound expression. EvoGAN uses an EA to search target results in the data distribution learned by GAN. Specifically, we use the Facial Action Coding System (FACS) as the encoding of an EA and use a pre-trained GAN to generate human facial images, and then use a pre-trained classifier to recognize the expression composition of the synthesized images as the fitness function to guide the search of the EA. Combined random searching algorithm, various images with the target expression can be easily sythesized. Quantitative and Qualitative results are presented on several compound expressions, and the experimental results demonstrate the feasibility and the potential of EvoGAN. The source code is available at https://github.com/faceeyes/EvoGAN.
\end{abstract}


\begin{highlights}
\item Current facial expression synthesis technologies cannot generate compound expressions precisely and gradient-based methods loss the diversity while generating.
\item The problem of generating various compound expressions can be formulated into an optimization problem.
\item We propose EvoGAN, which uses evolutionary algorithms, GANs and facial expression recognizers together to generate various arbitrary compound expressions.
\item EvoGAN models every input figures and generates an accurate expression according to its own features.
\end{highlights}

\end{frontmatter}

\section{Introduction}

\begin{figure*}[!h] 
\centering 
\includegraphics[width=\textwidth]{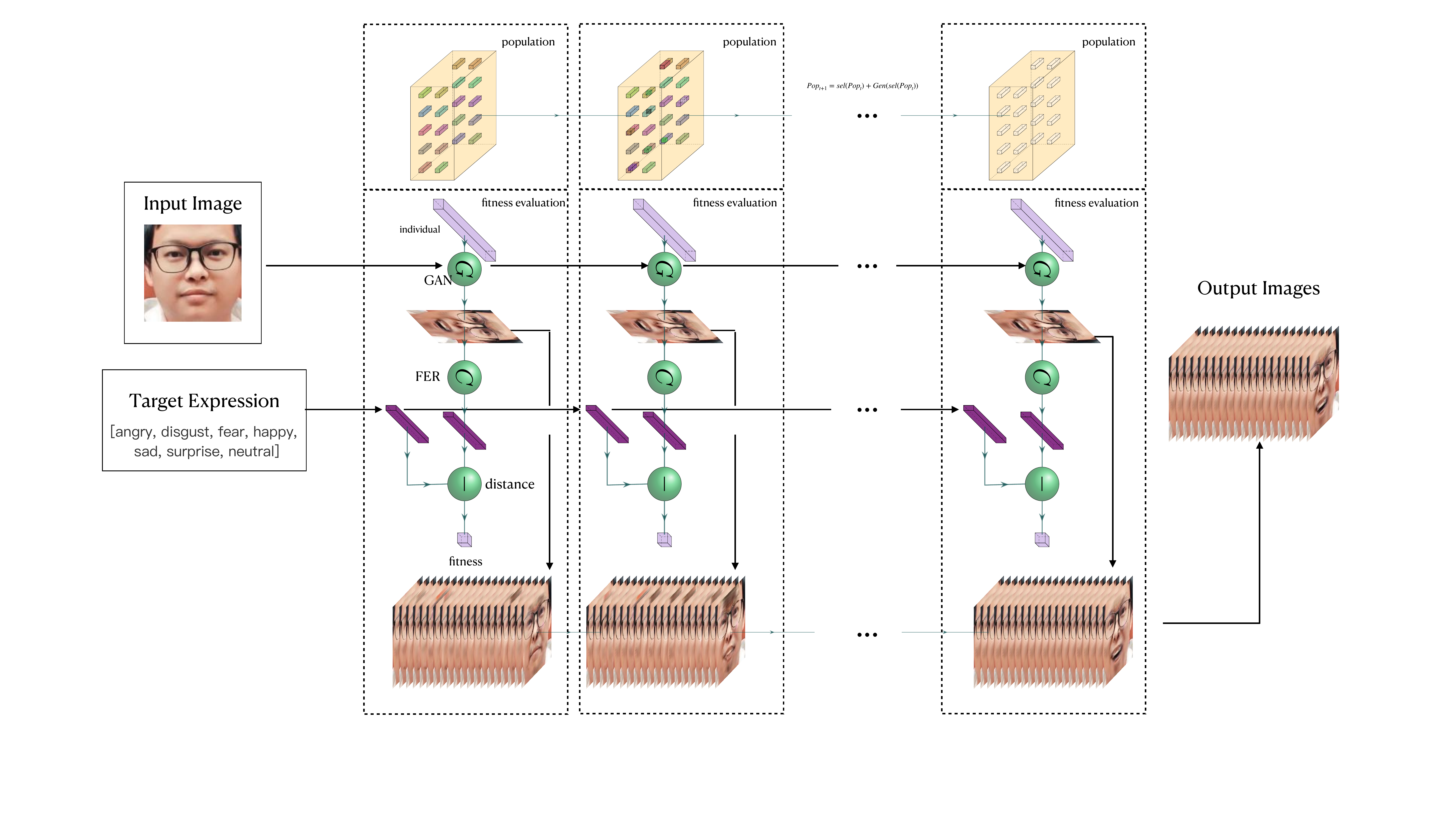}
\caption{The architecture of EvoGAN. The input of EvoGAN is an image and a seven-length array. The big boxes above shows the evolution process of EA. According to the fitness of every individual, the selection, mutation and crossover are conducted and breeding the next generation with a mixed gene. The middle part of the image shows the evaluating process. Every individual and the original image are input of the GAN and an image with expression is generated. A facial expression recognizer evaluates the expression of the generated image, and then the distance between the target expression and it is calculated, which is the final fitness score of the individual. During the process, the image generated by GAN is saved. When the fitness of the population reaches our target, all the images would be the output.}
\end{figure*} 

Generating images has a wide application in various fields including cinematography, graphic design, video games, medical imaging, virtual communication and ML research\cite{shoshan2021gancontrol}. Generative Adversarial Networks(GAN) plays a minimax game between a generator network and a discriminator network\cite{NIPS2014_5ca3e9b1}. The two networks are trained simultaneously, the generator tries to fake data while the discriminator tries to distinguish the fake ones and real ones. This process enables GAN to learn the data distribution of the dataset when the discriminator cannot tell which is real. However, the input of GAN is some random noise, and GAN map these random data with the target so that sometimes we may not be able to control the target accurately. Conditional GAN is developed to generate output conditioned on class labels. Deep Convolutional Generative Adversarial Network(DCGAN)\cite{radford2016unsupervised} combines both convolutional neural networks (CNNs) and GAN, allowing GANs to generate photorealistic images. With these ideas, recent works have made impressive breakthroughs in generating controllable face images.

Evolutionary Algorithm(EA) was first proposed by \cite{1967The} with the idea of natural evolution and was developed by \cite{holland1992adaptation}. EA has a great proficiency in solving optimization problems and has made great progress in economics, mathematics, geology, industry, social sciences, bioinformatics, etc. The fittest survives. EA uses a fitness function to evaluate the goodness of the solutions, and then, through selection, eliminates the bad solutions and retain good ones. With crossover and mutation, EA does not easily trap into locally optimal solutions and is able to search for the better ones. More importantly, EA does not require any mathematical features of the target function, meaning it does not come with any problems like gradient extinction or gradient explosion.

GAN is powerful in learning the data distribution. However, the existing facial expression datasets only have class labels instead of an accurate number, making it hard to generate accurate compound expressions.  There are generally two kinds of ways to generate facial expression images. One uses class labels like happy or sad as the input to generate the facial expressions and cannot generate compound expressions. The other imitates the facial movements from another image, which is completely different from generating an expression as the feature of  an individual also contributes to the expression. Further more, based on gradient descent, these methods only have one fixed output with one input. In fact, human can use different facial movements to express the same emotion. 

Since GAN\cite{NIPS2014_5ca3e9b1} first generated some black and white low resolution facial images, the capabilities of GANs have progressed quickly\cite{10.1145/3422622}. Besides the impressive improvements of the image quality, StarGAN\cite{Choi_2018_CVPR}  made a great progress changing the image-to-image translation type from single domain to multi-domain. GANimation\cite{pumarola2018ganimation} introduces FACS to facial expression synthesis and uses an array instead of a label as the input to make the synthesis much more flexible. \cite{shoshan2021gancontrol}presents a novel framework for training GANs with explicit control over generated images, which is extendable beyond the domain of human faces unlike the traditional CGAN methods. For some applications like face swapping or bringing old photos alive, these works have made great contributions. However, from the perspective of affective computing, we are not satisfied with the current status that the generated expressions are basic emotions and they do not have any variations. Moreover, from the perspective of evolutionary algorithms, researchers have been discovering more fields that are more related to application instead of mathematical problems. Quite a few works explored ways to both optimizing GANs using EAs and optimizing EAs using GANs\cite{he2020evolutionary, costa2020exploring, wang2019evolutionary, 10.1145/3377929.3398138}. EvoGAN is also an attempt to apply evolutionary algorithms in some applications.

In this paper, we aim to propose a framework using EA to search for the target within the data distribution learned by GAN. With this framework, we develop a model that generates compound expressions. In this model, we do not try to learn the data distribution of every expression but learn how to synthesize facial images with any facial movements, and then use EA to search the target expression. EA uses Facial Action Coding System (FACS), which describes facial expressions with the contractions of specific facial muscles, as the encoding. As shown in Figure 1, A facial expression generator is used to transfer the individuals in EA to an image. Then we use a facial expression recognizer(FER) to evaluate the expression of that image. Finally the distance between the expression of the synthesized image and the target expression will be the evaluation of the fitness of the individual. Through selection, crossover and mutation, EA can converge the result on the target expression. As a group random searching algorithm, EA maintains a group of good solutions, keeping the diversity of the output images. Quantitative results show that our method provides better results compared with generating images with a linear composition of two expressions. Quantitative results shows the convergence of EA, proving the feasibility of our method. Moreover, our framework does not depend on any of the parts of it, which means with the further development of GAN in the field of facial expression editing, our framework still works and can generate better images and with the further development of FER, our framework can generate expressions more accurately.

The contributions of this paper are summarized as follows:
\begin{itemize}
\item We first propose a novel general framework that combines EA and GAN to work as a whole.
\item We propose EvoGAN that generates face images with compound expressions with any combination of basic expressions based on the framework.
\item EvoGAN can generate target images with good diversity, which is more close to the reality and has the potential to make datasets.
\end{itemize}

\section{Related Work}

\textbf{Generative Adversarial Network.} GAN\cite{NIPS2014_5ca3e9b1} is proved to be powerful in learning the data distribution. GAN is called adversarial because it has a generator and a discriminator and plays a game between them. Typically, the generator and the discriminator are both neural networks and are trained simultaneously. The generator generates fake samples while the discriminator discriminates the fake samples and the real data. After the process of adversarial training, since the generator has already learned the probability distribution of the real data, the discriminator cannot tell them apart. \cite{radford2016unsupervised} proposed DCGAN introducing CNNs to GAN, enabling to learn a hierarchy of representations from object parts to scenes. With DCGAN, high-resolution and realistic anime avatars can be generated. Plenty of work has been done to advance the ability of GAN. \cite{Mao_2017_ICCV} proposed Least Squares GAN to overcome the vanishing gradients problem. \cite{miyato2018spectral} proposed spectrally normalized GANs to overcome the instability of the training of GANs. \cite{pmlr-v97-zhang19d} proposed Self-Attention GAN allows attention-driven, long-range dependency modeling for image generation tasks. GANs have a wide application in super-resolution imaging\cite{Wang_2018_ECCV_Workshops,Ledig_2017_CVPR}, classification\cite{9125995}, image synthesis\cite{pmlr-v119-golany20a}, color mapping\cite{9047180}and face generation \cite{karras2018progressive,radford2016unsupervised}.

\textbf{Facial expression editing.} Facial expressions editing is challenging as it requires high-level understanding of input facial images and prior knowledge about human expressions\cite{Wu_2020_CVPR}. When generating face images, conditional GANs are used so that the attributes of the output image can be controlled. \cite{perarnau2016invertible} uses encoders to inverse the mapping of a cGAN, mapping real images into a latent space and a conditional representation, named IcGAN. With IcGAN, facial images with specific attributes are first generated. After that, some works have been done to improve the performance\cite{li2016deep, zhu2020unpaired, he2019attgan}.\cite{Choi_2018_CVPR} proposed StarGAN, which is a unified GAN for multi-domain image-to-image translation, enabling to control hair color, gender, age, skin color, expressions, etc. of the input image at the same time.\cite{liu2019stgan} proposed STGAN for arbitrary image attribute Editing.\cite{pumarola2018ganimation} proposed GANimation, which is based on FACS, overcoming the problem that GAN can only generate discrete number of expressions .\cite{Wu_2020_CVPR} proposed Cascade EF-GAN, a novel network that performs progressive facial expression editing with local expression focuses, which generates images with much less artifacts and blurs. \cite{shoshan2021gancontrol} proposed a framework for training GANs so that control is not constrained to morphable 3D face model parameters and is extendable beyond the domain of human faces.

\section{Problem Formulation}

We define an RGB image as $I \in R^{H\times W\times 3}$, representing the color of each pixel of the image. A set of N action units $y=(y_1,...,y_N)^\top$, where each one represents the intensity of one specific, single movement of face while the whole set of action units represents a complete gesture expression. Moreover, we use an array 
\begin{equation}
e = [e_{angry},e_{disgust},e_{fear},e_{happy},e_{sad},e_{surprise},e_{neutral}]
\end{equation}
to represent the evaluation of expression combined of eight basic expressions. 
Our aim is to learn a mapping $M$ to translate an input image $I$ into an output image $I_{y}$conditioned on an expression target $e_t$.  Since current GANs cannot realize this target, we use GANs that use Action Units $y$ as input. This kind of GANs can synthesize arbitrary facial images. As soon as we find out the correct $y$ that indicates the facial movements of the expression we want, we can use GAN to synthesize the expression. 
\begin{equation}
I_t = GAN(I, y_t)
\end{equation}
Moreover, for every single face, they do not have exactly the same facial movements to show the same expression, which makes it important to calculate $y$ for every face instead of transfer.To realize this, we use evolutionary algorithms to search for the target $y$ and transfer the problem into an optimization problem minimizing the distance between the current synthesized expression and the target expression. The formulation is:
\begin{equation}
min f(y)=distance(g(I, y), e_t)
\end{equation}
Then, the action units array that minimizes $f(y)$ is the target $y_t$we want. With this $y_t$, we can finally get our target facial expression image.

\section{EvoGAN}

\subsection{EvoGAN Architecture}

The overall architecture of EvoGAN is shown in Figure 1. The blocks above show the procedure of evolution that the individuals of the population would evolve through crossover and mutation and finally having the best fitness to the environment. The ones in the middle show the procedure of evaluating the fitness of an individual. GAN generates an image with the input image and the evaluating individual which denotes the gesture expression. After that, the image is evaluated by a pre-trained VGG19 facial expression recognizer with an output. Finally, the distance between the output of VGG19 and the target expression is calculated and sent back to EA as the fitness of the individual. 

EA is mostly used in solving mathematical problems, EvoGAN uses AU as the key connector between EA and GAN. The input of GAN is exactly the individuals in EA. So, the individuals in EA are seemingly arrays but actually different facial images. Moreover, EA needs to judge all the individuals to decide whether they can survive or not. Having GAN already transformed the individuals into images, we use a pre-trained deep neural network to recognize the expression of the images. The result can be easily turned into the final score of the individual, driving EA to move on the evolve process.

\subsection{Evolutionary Algorithm}

\begin{figure}[!htbp] 
\centering 
\includegraphics[width=4cm]{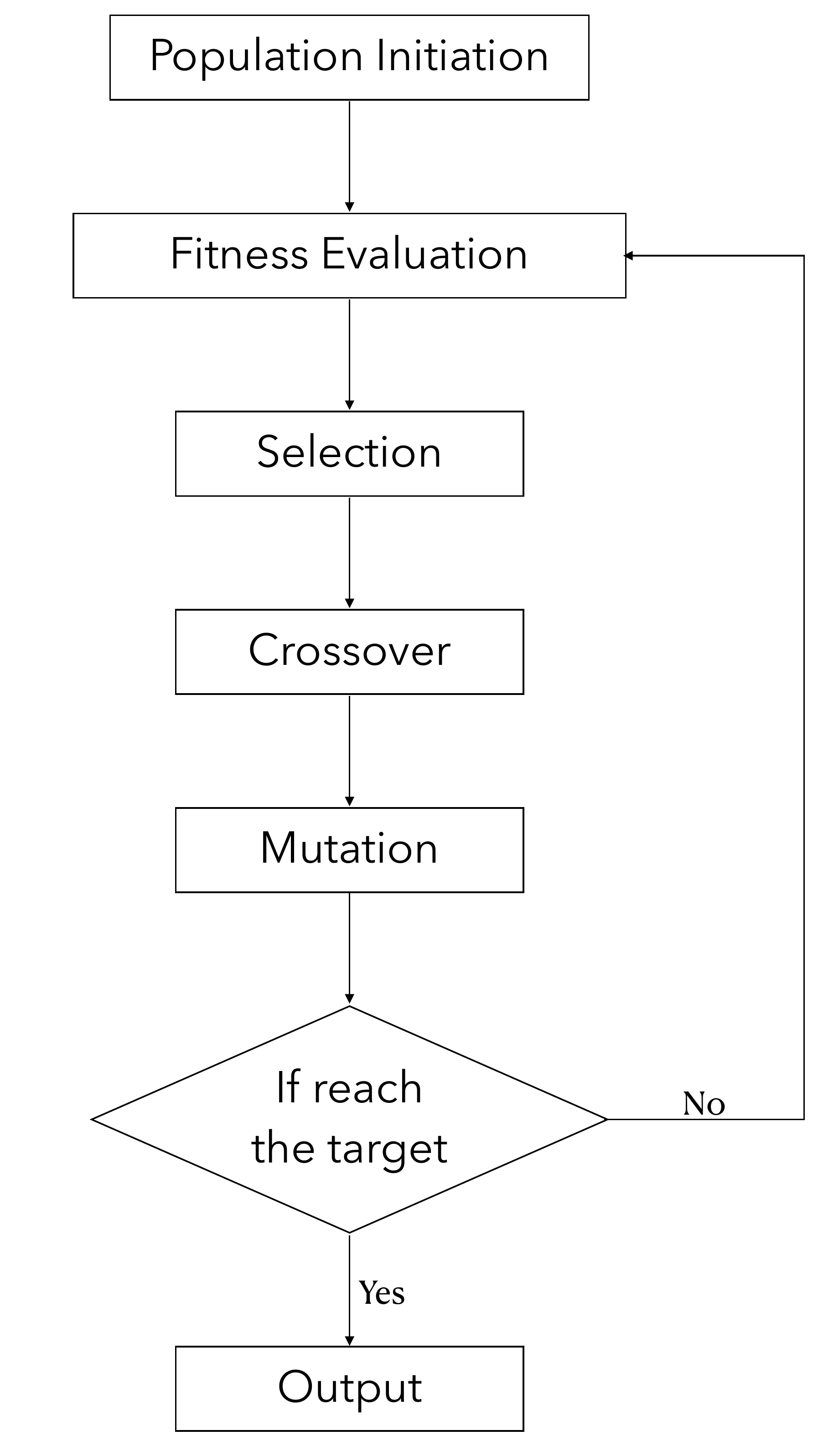}
\caption{The architecture of EA. The modules and parameters are designed based on the specific problem.}
\end{figure} 

EA is a kind of methodology dealing with optimization problems. With the idea of biological evolution, EA maintains a population and calculates the fitness of all the individuals. The fittest survives and leaves its offspring through crossover and mutation. Generations later, the population fits the environment well. In all, EA is a global optimization algorithm based on random searches. Meanwhile, EA does not require the mathematical feature of the fitness function so that EA can still handle the problem when the fitness function consists of GAN and VGG.

As shown in Figure 2, A simple Genetic Algorithm(GA) comprises the following steps. First of all, a set of solutions, which are called individuals, are randomly generated as a population.
\begin{equation}
Pop_0 = initiation()
\end{equation}
After that, the population does crossover and mutation to search the target. 
\begin{equation}
Q_t = Gen(Pop_t)
\end{equation}
All the individuals have a probability of crossover and mutation. The higher probability of crossover indicates the population prefers to doing more local search and the higher probability of mutation indicates the population prefers to exploring. Generally, at the begin of the algorithm, it would be better to have more explore and at the end of the algorithm, it would be better to do more local search and converge. Thus, an adaptive strategy is used in order to realize that:
\begin{equation}
p_c = k_c * (f_{best} - f) / (f_{best} - f_{avg})
\end{equation}
\begin{equation}
p_m = k_m * (f_{best} - f) / (f_{best} - f_{avg})
\end{equation}

Specifically, $p_c$, $p_m$ indicates the probability of crossover and mutation, $k_c$, $k_m$ are hyperparameters that control the probabilities, $f_{best}$, $f_{avg}$ and $f$ indicates the fitness of the best in the population, the average and the current individual. When the population is going to converge, $f_{best} - f_{avg}$ becomes smaller. If the current individual is close to the best individual, it would be better to keep it the unchanged.

The goodness of these solutions are evaluated by the fitness function. Based on the fitness of each individual, the better ones would survive while the ones that do not fit the environment well are eliminated. 

\begin{equation}
Pop_{t+1} = sel(Pop_t + Q_t)
\end{equation}
In EvoGAN, the situation is different from traditional EAs. Generally, optimal questions only focus at the best solution, but EvoGAN hopes for a variety of good solutions. Generally, the selection function is based on probabilities. The roulette wheel selection selects every one of the populations at the probability of $f_i/\sigma f_j$ until the next generation reaches the maximum number. In EvoGAN, these kinds of selection functions may leads to premature and loss of variety. Specifically, the whole population gradually becomes the copy of one individual. Moreover, considering the purpose of EvoGAN is generating various images with one target expression, a sort selection function is used. 

\begin{equation}
Pop_{t+1} = sort(Pop_t + Q_t)[0:p - 1]
\end{equation}

Where $p$ indicates the maximum number of the population. This means the next generation selects the best $p$ individuals in $Pop_t$ and $Q_t$, which maintains the variety well and also converges.

After that, the next population does crossover and mutation.The whole process is repeated until the optimal solution is found. 

\subsection{EvoGAN}
In our model, we do not care much about the speed of the convergence, as a result, using some basic methods in initiation, selection, crossover and mutation would be sufficient. To initiate the population, we generate $N$ individuals $I_n$ with $M$ random float numbers ranging from 0 to 1. After evaluating the fitness of each individual $f_{I_n} = f(I_n)$, we use the roulette method to select $\frac{N}{2}$ individuals by a chance of $\frac{f_{I_n}}{\sum_{i=1}^N{f_{I_i}}}$, and to maintain convergence, the fittest individual always survives. Then, for each individual $I_a$, it has a chance of $P_c$ to cross over with another individual $I_b$. If happens, two random number $i, j$ between 0 to $M - 1$ would be generated, and $I_a[i:j]$ would be swapped with $I_b[i:j]$. Moreover, it also has a chance of $P_m$ to mutate. If happens, a random number $k$ between 0 to $M - 1$ would be generated, and $I_a[i]$ would have a half-half chance to be doubled or halved. The algorithm stops when the fittest individual reaches the fitness of $f_{max} - e$.

To evaluate an individual $ y=(y_1,...,y_N)^\top$, we use a GAN to translate input $ y $ and image $ I $ to image$ I_{y} $. 
\begin{equation}
I_{y}=GAN(I, y)
\end{equation}
Then a pre-trained VGG19 facial expression recognizer is used to evaluate the expression of  $ I_{y} $. 
\begin{equation}
e=FER(I_{y})
\end{equation}
Finally the fitness function is the distance between $ e $ and target expression $ e_t $. 
\begin{equation}
\begin{split}
f(y)=distance(e, e_t)=distance(FER(I_{y},e_t)\\
=distance(FER(GAN(I, y)),e_t)
\end{split}
\end{equation}

\begin{algorithm}[!htb]
\caption{EvoGAN}
\label{alg:algorithm}
\textbf{Input}: An Image X, a target expression Et.\\
\textbf{Parameter}: Population size S, crossover probability Pc, mutation probability Pm.\\
\textbf{Output}: A set of images Y with target expression.\\
\begin{algorithmic}[1]
\STATE randomly generate S individuals.
\FOR{t = 1 to T}
\FOR {each individual I in population}
\STATE image = GAN(X, I)
\STATE expression = FER(image)
\STATE I.fitness = distance(expression, E)
\ENDFOR
\STATE selection();
\FOR {each individual I in population}
\STATE crossover if Pc;
\STATE mutation if Pm;
\ENDFOR
\ENDFOR
\STATE \textbf{return} solution
\end{algorithmic}
\end{algorithm}

\begin{figure}[htbp] 
\centering 
\includegraphics[height=\textheight]{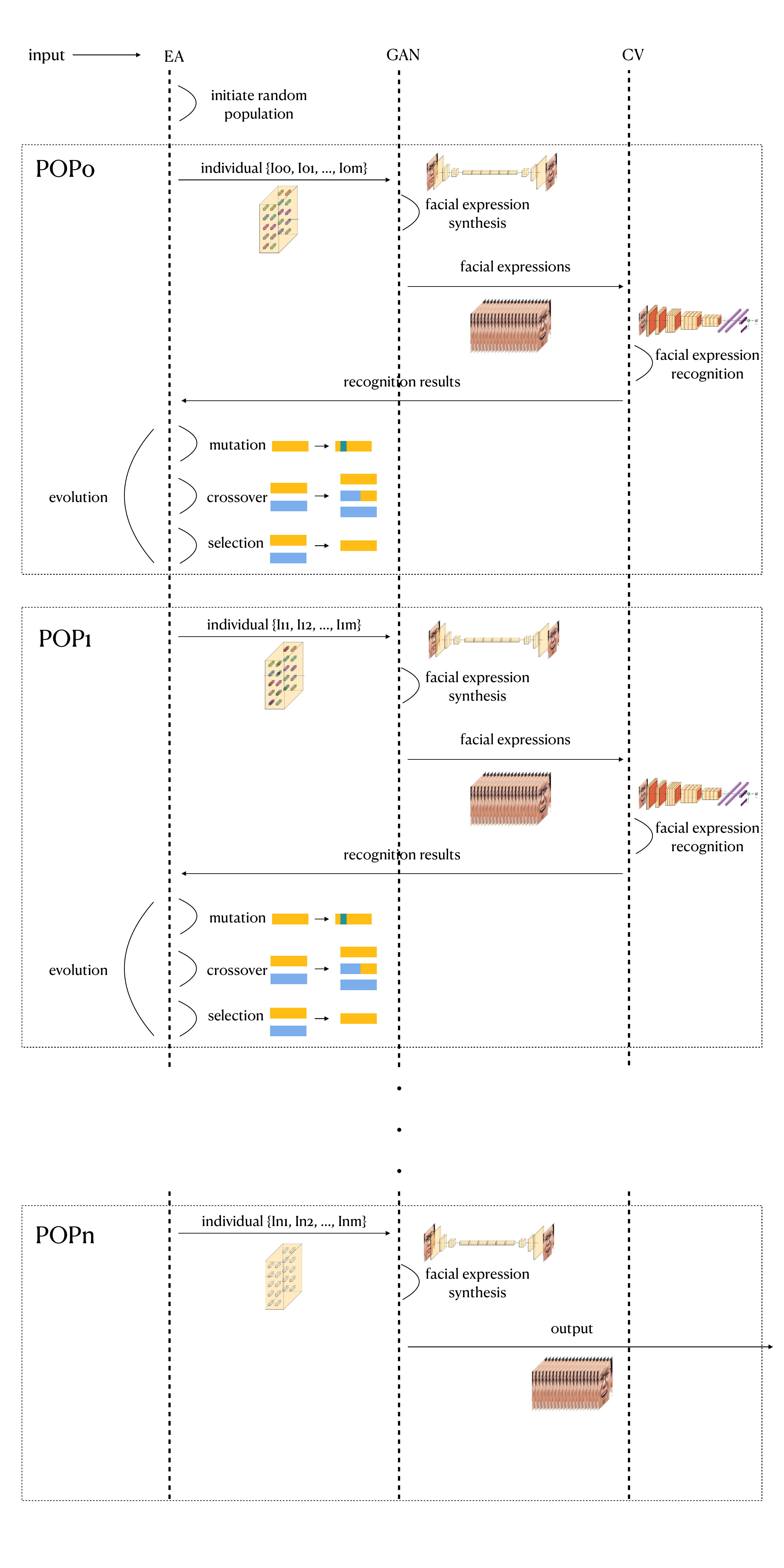}
\caption{The sequence diagram of EvoGAN.}
\end{figure} 

Figure 3 shows how EvoGAN works with a sequence diagram. First of all, the target expression $e_t$ and the source image are put into EvoGAN. After EA initializes the first population, it judges all the score by transforming all the individuals to images with GAN and then using deep neural network to recognize the expression of the images. After that, all the individuals get the recognition results as their fitness scores. So, the "fitness evaluation" part of EA is done and then is the ordinary selection, mutation, crossover process. With all these procedures, EA optimizes the facial movements of the input facial images according to the expression recognition of this set of facial movements. For example, an individual with AU12 which means "Lip corner puller" may get a higher score when the target expression is "happy". Such individual with a higher score has a higher possibility to crossover and to be selected, thus its "Lip corner puller" gene will be preserved well with the evolution. Those AUs that are not related to "happy" have a lower chance to survive. Finally, the survivals mostly have higher numbers in AU6 "Cheek raiser", AU12 "Lip corner puller", etc. which can get higher scores during expression recognition.

Using EA, EvoGAN not only can optimize the images to an accurate expression, but also preserves a population of images with the target expression. These images has different facial movements but are all optimized to the target expression.

\section{Experiments}
\subsection{Implementation}

For EA, the population size is 50 and the generation number is 50. The basic probability of crossover is 0.7 and mutation is 0.02. For GAN and FER, we use pre-trained GANimation and VGG19. Since there is not any training process, the whole experiment is conducted on macOS.

\subsection{Quantitative Evaluation}
Basically the final output is consist of four aspects: image quality, time cost, the goodness of the solution, diversity. The quality of the images are determined by GAN model and FER model. GAN determines the visual quality, including artifacts and blurs. FER determines the expression quality. Specifically, if EvoGAN has already converged to the target expression, but human consider the output image far from the target, it is caused by an undesirable facial expression recognizer. While the specific models of GAN and FER are not the key point of our proposal, the image quality is paid less attention. 

The other three aspects that show the feasibility and potential of the EvoGAN framework, are determined by EA. Like other optimal questions, it is easy to understand that the time cost indicates the computational burden and the goodness of the solution indicates the degree of optimization. If the population size is set 1, EA becomes a local search algorithm, and if the population size is set infinity, EA becomes a completely random algorithm. Thus, a small population size may limit the goodness of the solution while a big population size leads to a large time cost. The generation number balances the time cost and the goodness of the solution. The basic probability of crossover and mutation affects both the time cost and the goodness of solution. Crossover is similar to local search while mutation is similar to a limited global random search. A proper probability of crossover keeps the algorithm to converge and a proper probability of mutation prevents the algorithm from a local optimal solution. Meanwhile, the design of initiation, crossover, mutation, selection functions also affects the time cost and the goodness of the solution.

Except the parameters and the algorithm, the input also have a significant impact on the results. Figure 4 shows the output with different input images and same facial movements. It can be inferred that the facial identity features affects the expression of emotion and the recognition results. Thus, when doing the searching, EvoGAN actually does the model for only the input image instead of building a model for everyone. Compared with the identity features, the target expression has much more impact on the results. An experiment is conducted using random AU factors and GANimation model, and the generated results are recognized and averaged. The average output is shown in Figure 5. From Figure 5, we can observe that "happy" occupies a great proportion in the search space while the others occupies about 20 percent of the search space. This indicates that, for example, searching for an expression with "happy" would be definitely easier than searching for an expression with "surprised". As different expression occupies different proportion in the search space, different target expression has different time cost.

\begin{figure}[htbp]
\centering
\begin{minipage}[t]{0.45\textwidth}
\centering
\includegraphics[width=\textwidth]{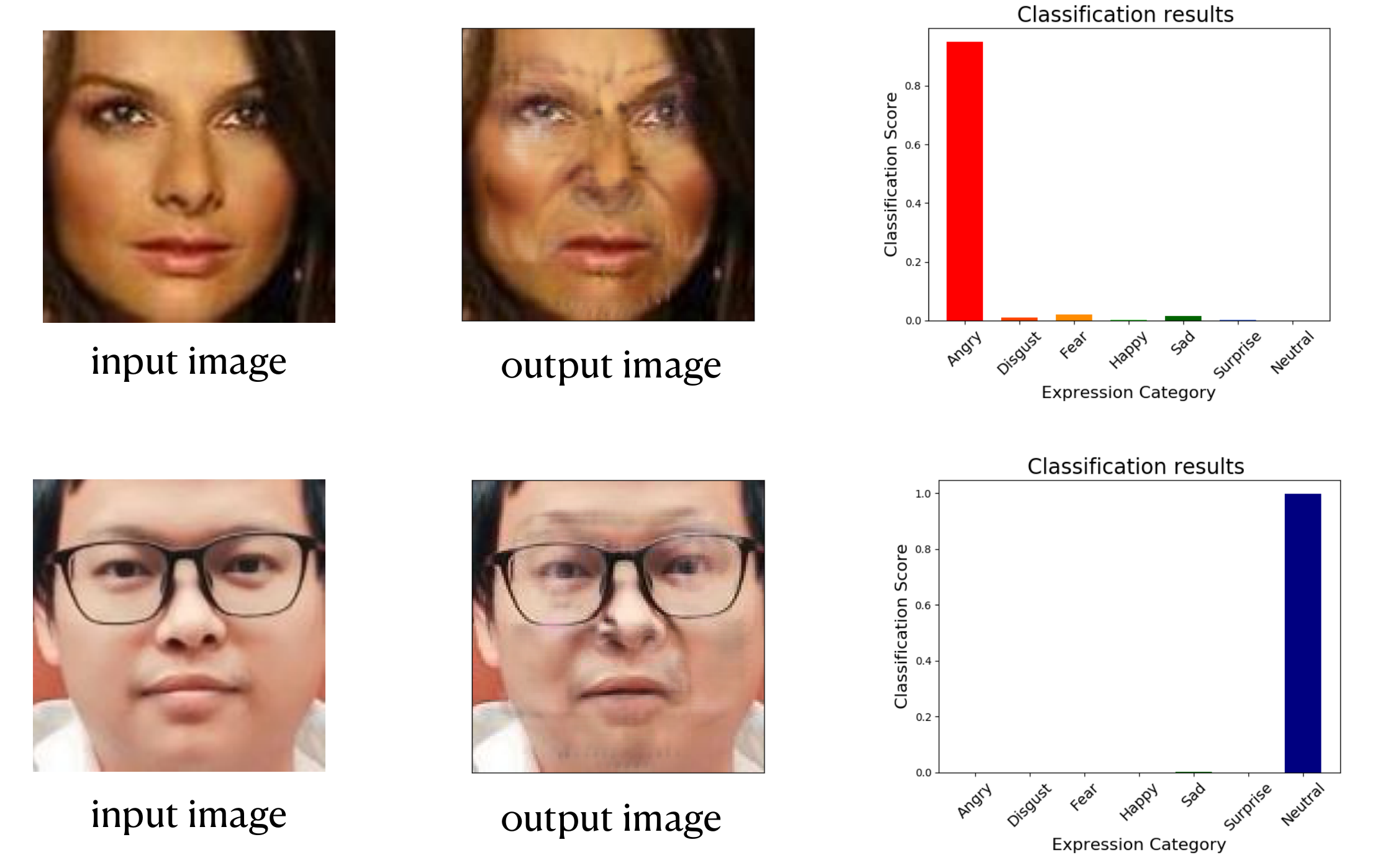}
\caption{The output with different input images and the same facial movements.}
\end{minipage}
\begin{minipage}[t]{0.45\textwidth}
\centering
\includegraphics[width=6cm]{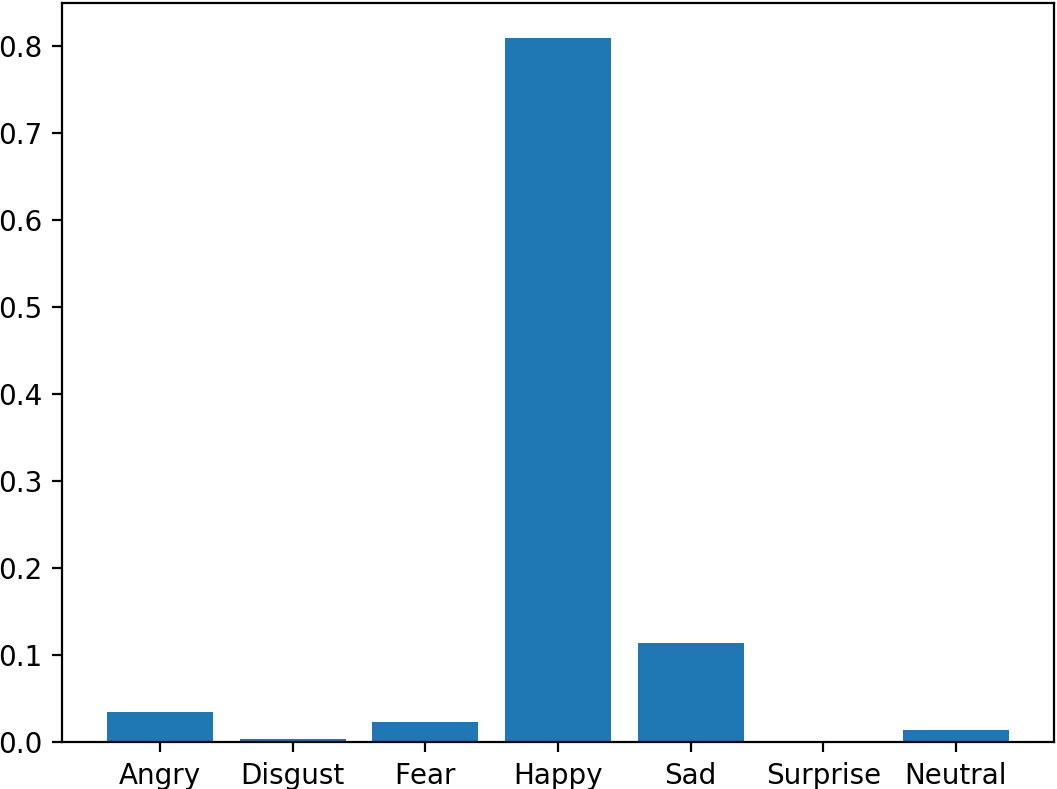}
\caption{The average output expression with random facial movements.}
\end{minipage}
\end{figure}

Figure 6 shows the statistical data of EA with different sizes of population and generation. The horizontal coordinate represents the generation while the vertical coordinate represents the score of every individual. Every point in Figure 6 represents an individual. It can be seen that with a larger population size, the algorithm is more likely to converge earlier. However, the time cost can be easily estimated by multiplying generation count and population size. With 2,500 times generating and estimating, the experiment with generation 50 and population size 50 costs more time than the one with generation 80 and population size 30. To make sure the algorithm converges smoothly, neglecting the time cost, the generation size and the population size are all set 50.

\begin{figure}[htbp]
\centering
\begin{minipage}[t]{0.45\textwidth}
\centering
\includegraphics[width=\textwidth]{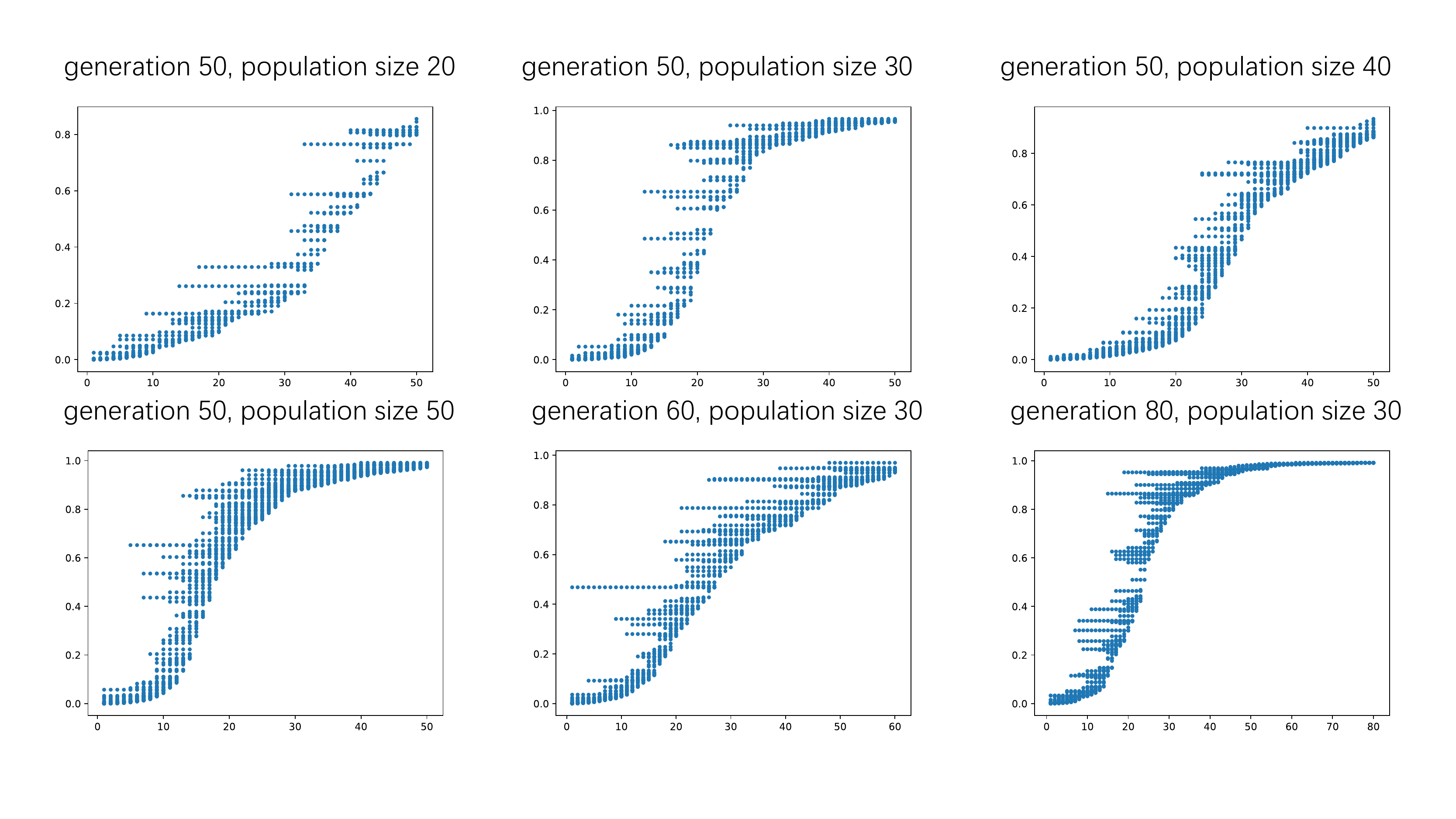}
\caption{The statistical data of EA with different sizes of population and generation.}
\end{minipage}
\begin{minipage}[t]{0.45\textwidth}
\centering
\includegraphics[width=\textwidth]{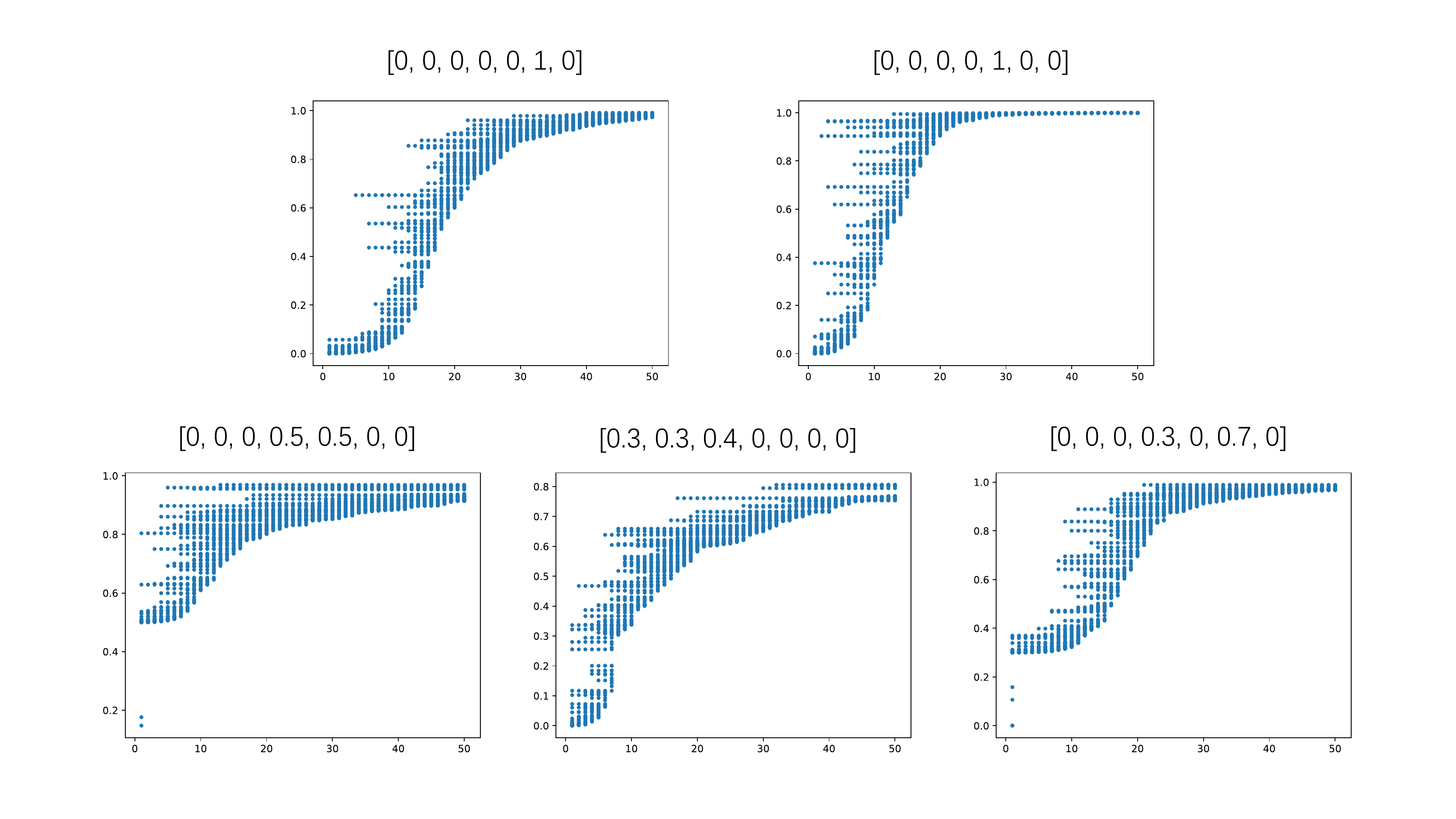}
\caption{The statistical data of EA with different target expressions. }
\end{minipage}
\end{figure}

Figure 7 shows the statistical data of EA with different target expressions. The array represents the target expression in the order of anger, disgust, fear, happy, sad, surprise, neutral. It can be observed that the ones with expression "happy" and "sad" converges much earlier and faster while the others take more time to converge. Experiments show that compound expressions also take more time to converge.

\subsection{Qualatative Evaluation}

Figure 8 shows 20 of the final output of EvoGAN with five target expressions. Due to the limitation of GANimation, there are still some artifacts and blurs in the images. While this paper focus on the generation of compound expressions and the diversity of output images, the image quality are not the key point. From Figure 8, we can observe that the diversity of the results is maintained to some extent. On the other hand, since the recognition result of these images has already reached the target expression, the performance of EvoGAN is limited by the performance of FER.

\begin{figure}[htbp] 
\centering 
\includegraphics[width=\textwidth]{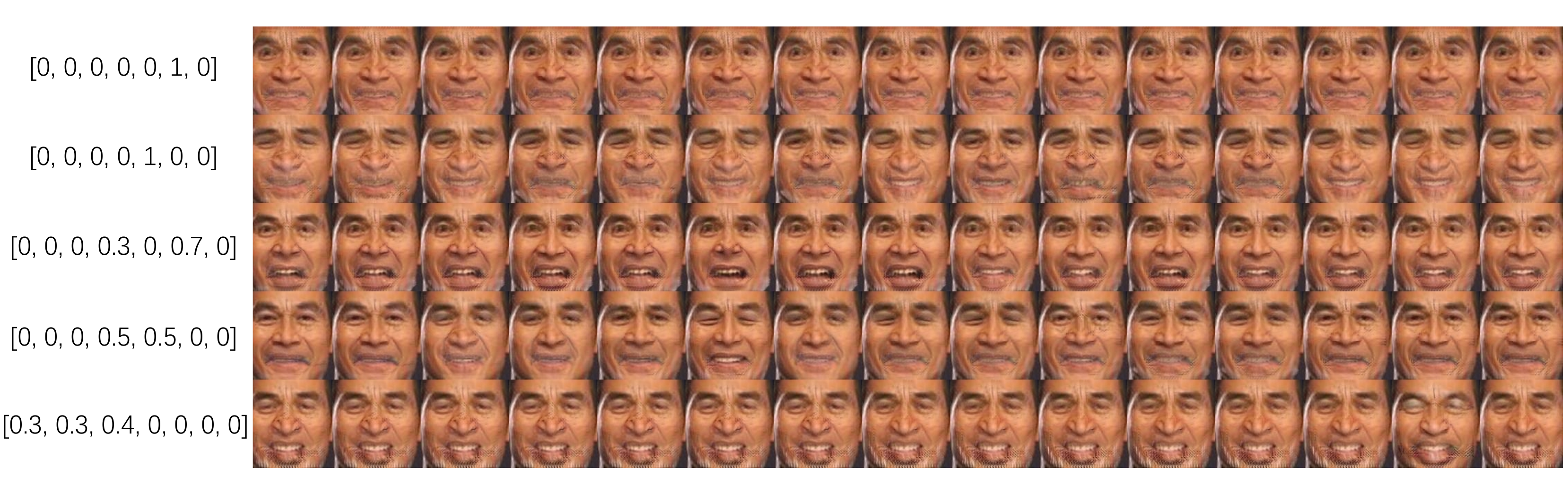}
\caption{The final results of EvoGAN. The target expression of first row is surprise: 1.0. The second row is sad: 1.0. The third row is happy: 0.3, surprise: 0.7. The fourth row is happy: 0.5, sad: 0.5. The fifth row is anger: 0.3, disgust: 0.3, fear: 0.4.}
\end{figure} 

Figure 9 shows the process results of EvoGAN with different target expressions. Five of the population are shown in Figure 9. It is obvious that in generation 0, the expressions are completely a mess. Then, the expressions gradually converge to the final output in Figure 8.

\begin{figure}[htbp] 
\centering 
\includegraphics[width=\textwidth]{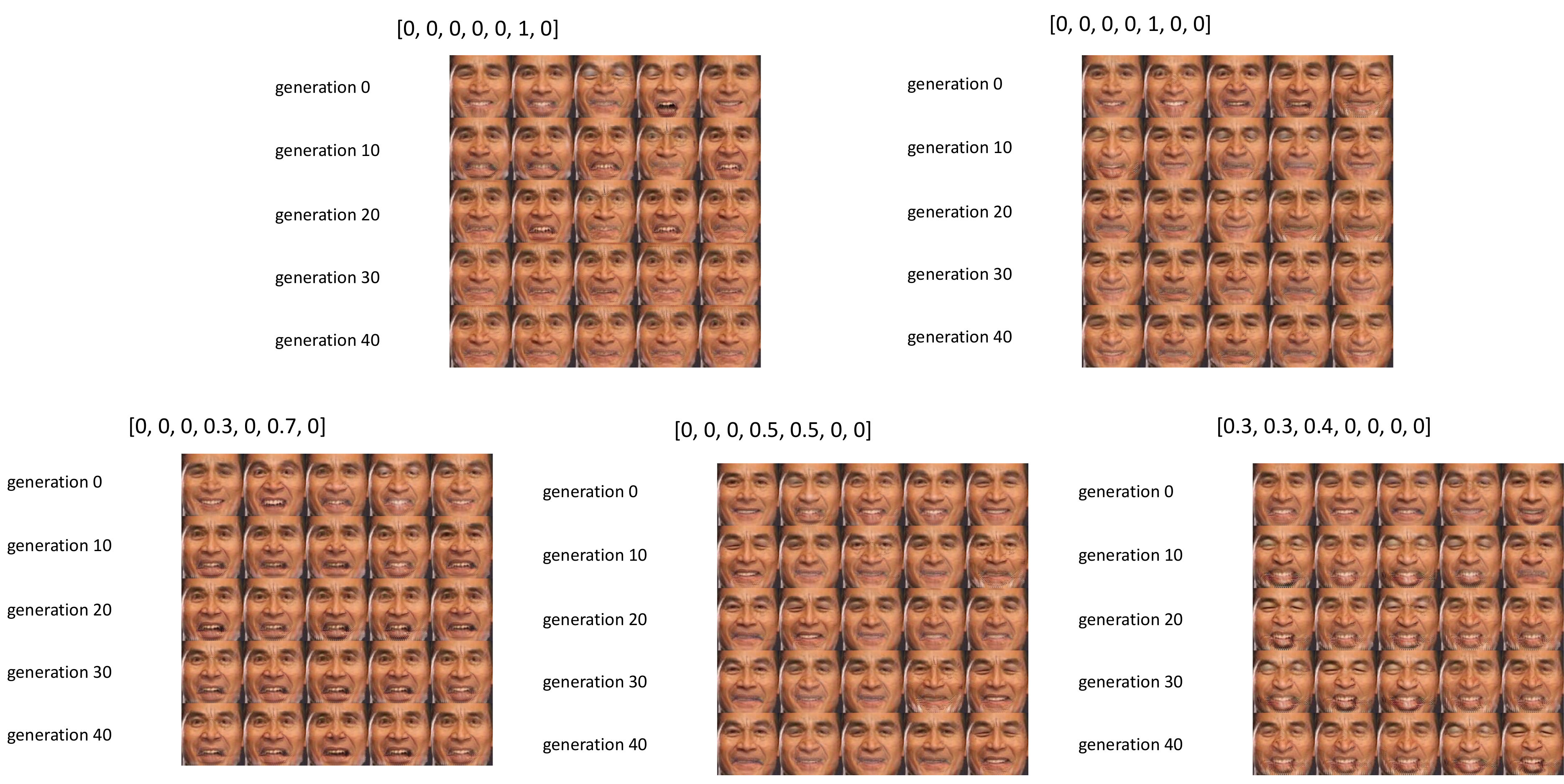}
\caption{The process results of EvoGAN with different target expressions. }
\end{figure} 

As the main purpose and contribution of EvoGAN is quite different from the related methods. Table 1 compares EvoGAN with other related methods to show that difference. From IcGAN\cite{perarnau2016invertible} to CycleGAN\cite{Zhu_2017_ICCV}, plenty of works have been done to improve the image quality of the output images. StarGAN\cite{Choi_2018_CVPR}  made a great progress changing the image-to-image translation type from single domain to multi-domain and further improve the image quality. GANimation\cite{pumarola2018ganimation} introduces FACS to facial expression synthesis and uses an array instead of a label as the input to make the synthesis much more flexible. Cascade EF-GAN\cite{Wu_2020_CVPR} improved the image quality of GANimation. EvoGAN is implemented based on GANimation, having the same image quality. EvoGAN focuses on generating arbitrary compound expressions using an array and using random search algorithm to maintain the diversity of the output.

\begin{table}
\centering
\caption{Comparison with related methods.}
\resizebox{.95\columnwidth}{!}{
\begin{tabular}{c|cc|cc|ccc|cc}
\toprule
\multirow{2}{*}{method}  & \multicolumn{2}{c|} {translation type} & \multicolumn{2}{c|} {input type} & \multicolumn{3}{c|} {expression type} & \multicolumn{2}{c} {optimization strategy} \\
\cline{2-10}
& single domain translation & multi-domain translation & label & array & basic expressions & mimc expression & compound expression & gradient & random search\\
\midrule
IcGAN\cite{perarnau2016invertible}                  & $\checkmark$ &  & $\checkmark$ &  & $\checkmark$ &  &  & $\checkmark$ &  \\
DIAT\cite{li2016deep}                      & $\checkmark$ &  & $\checkmark$ &  & $\checkmark$ &  &  & $\checkmark$ &  \\
CycleGAN\cite{Zhu_2017_ICCV}             & $\checkmark$ &  & $\checkmark$ &  & $\checkmark$ &  &  & $\checkmark$ &  \\
StarGAN\cite{Choi_2018_CVPR}                &  & $\checkmark$ & $\checkmark$ &  & $\checkmark$ &  &  & $\checkmark$ &  \\
GANimation\cite{pumarola2018ganimation}           &  & $\checkmark$ &  & $\checkmark$ &  & $\checkmark$ &  &  $\checkmark$ &  \\
Cascade EF-GAN\cite{Wu_2020_CVPR}  &  & $\checkmark$ &  & $\checkmark$ &  & $\checkmark$ &  &  $\checkmark$ &  \\
EvoGAN(Ours)      &  & $\checkmark$ &  & $\checkmark$ &  &  & $\checkmark$ &  & $\checkmark$ \\
\bottomrule
\end{tabular}}
\end{table}

\section{Conclusion and Future Work}

In this work, we propose an evolutionary algorithm assisted GAN framework called EvoGAN to generate various compound expressions. Specifically, we transfer the compound expression synthesis into an optimization problem and use EA to search for the optimal. GAN and FER are used as the fitness function to evaluate the individuals in EA population. The experimental results demonstrate the feasibility and the potential of EvoGAN. Moreover, all our experiments are performed based on GANimation and VGG-19, which are not sort-of-the-art GANs or FERs and they do not fits our model well. EvoGAN framework can also be embedded by many other GAN and FER variants and some changes can be made to advance the whole model. 

However, since the initial purpose of this paper is applying to some psychological researches or making compound expression datasets, in applications that more close to daily life, it might be sufficient to use models like GANimation instead of EvoGAN. Moreover, compared with these models, though EvoGAN can use pre-trained models, EvoGAN costs pretty more time than these models when generating images. For example, If the population size is 50 and generation is 50, EvoGAN would generate 2500 images using GANimation in all. But for doing researches, we can afford to spend more time to generate compound expression and to maintain diversity.

Our future work includes:
\begin{itemize}
\item Reduce the time cost of EvoGAN.
\item Replacing GANimation and VGG-19 with networks that better fit the whole framework and redesign EA for this problem.
\item Develop EvoGAN and generating compound expression datasets and micro-expression datasets.
\item Conducting psychological experiments based on the synthesized images.
\item Advancing compound expressions recognization based on the future research.
\end{itemize}

\section{Acknowledgement}
We thank the anonymous reviewers and our study participants for their time and helpful feedback. This work is supported by the Scientific and Technological Innovation 2030 Major Projects under Grant No. 2018AAA0100902, the Science and Technology Commission of Shanghai Municipality under Grant No.19511120600. Also supported by The Research Project of Shanghai Science and Technology Commission (20dz2260300) and The Fundamental Research Funds for the Central Universities.

\clearpage
\bibliographystyle{elsarticle-num} 
\bibliography{EvoGAN}

\end{document}